\documentclass[graybox]{svmult}

\usepackage{type1cm}        
\usepackage{makeidx}         
\usepackage{graphicx}        
\usepackage{multicol}        
\usepackage[bottom]{footmisc}
\usepackage{multirow}

\usepackage{newtxtext}       %
\usepackage{newtxmath}       

\makeindex             

\begin{document}

\title*{Are LLMs Robust for Spoken Dialogues?}

\author{Seyed Mahed Mousavi\textsuperscript{\textdagger}, Gabriel Roccabruna, Simone Alghisi\textsuperscript{\textdagger}, Massimo Rizzoli\textsuperscript{\textdagger}, Mirco Ravanelli, and Giuseppe Riccardi}

\institute{ \textsuperscript{\textdagger}Work partly conducted while on an internship at Mila - Quebec AI Institute. \\ \\
Seyed Mahed Mousavi, Gabriel Roccabruna, Simone Alghisi, Massimo Rizzoli, and Giuseppe Riccardi\\ Signals \& Interactive Systems Lab, University of Trento, Italy\\ \texttt{{mahed.mousavi,giuseppe.riccardi}@unitn.it}\\
\\ Mirco Ravanelli \\ Concordia University, Mila-Quebec AI Institute, Canada.\\
\texttt{mirco.ravanelli@mail.concordia.ca}\\ \\}

\maketitle

\abstract{Large Pre-Trained Language Models have demonstrated state-of-the-art performance in different downstream tasks, including dialogue state tracking and end-to-end response generation. Nevertheless, most of the publicly available datasets and benchmarks on task-oriented dialogues focus on written conversations. Consequently, the robustness of the developed models to spoken interactions is unknown. In this work, we have evaluated the performance of LLMs for spoken task-oriented dialogues on the DSTC11 test sets. Due to the lack of proper spoken dialogue datasets, we have automatically transcribed a development set of spoken dialogues with a state-of-the-art ASR engine. We have characterized the ASR-error types and their distributions and simulated these errors in a large dataset of dialogues. We report the intrinsic (perplexity) and extrinsic (human evaluation) performance of fine-tuned GPT-2 and T5 models in two subtasks of response generation and dialogue state tracking, respectively. The results show that LLMs are not robust to spoken noise by default, however, fine-tuning/training such models on a proper dataset of spoken TODs can result in a more robust performance.}


\section{Introduction}

Large Pre-Trained Language Models (LLMs) are Transformer-based \cite{attentionpaper} architectures with large numbers of parameters (pre-)trained in an unsupervised manner on a large amount of textual data. Recently, these models have outperformed other data-driven models and demonstrated state-of-the-art performance in open-domain response generation \cite{zhao-etal-2020-knowledge-grounded}, as well as task-oriented dialogue (TOD) modeling, including Dialogue State Tracking (DST) \cite{won2023break}, end-to-end response generation \cite{e2eres}, and joint DST and response generation \cite{hosseini2020simpletod, kulhanek2021augpt}. 

There have been several studies on the evaluations of LLMs as dialogue models demonstrating the limitations of such models including biased, toxic, and contradictory responses \cite{mousavi-etal-2023-response, limitsoftransfer}, as well as introducing false/unexisting information or irrelevant elements to the dialogue context \cite{humeval,zhang-etal-2020-dialogpt}. However, the robustness of these models to spoken dialogues is unknown. In the pre-training phase, LLMs are trained on huge amounts of written textual data from different sources including an unoptimized selection of books, news, social forums (as a distant substitute for dialogue data) \cite{zhang-etal-2020-dialogpt}, and Wikipedia webpages \cite{bert}. Further, when applied to the downstream tasks, the dialogue datasets and benchmarks used to fine-tune and evaluate LLMs are mostly focused on written dialogues. 

Spoken dialogues are different from written dialogues in terms of a) different expressions, vocabulary, and wording; b) the presence of possible speech disfluencies (such as pauses, segment repetitions, segment rephrasing, and speech error); and c) possible automatic speech recognition errors. Despite the abundance of written dialogue datasets, there is a lack of a proper dataset for spoken TODs for response generation and DST. There have been few attempts to address this problem in the context of recent Dialog System Technology Challenges (DSTC) \cite{kim2021robust,soltau2022speech}. These studies provided transcriptions of a small-scale dataset of spoken dialogues (mostly as a test set), and engaged the community to develop robust models for spoken interactions while using written dialogues for training/fine-tuning. The works participating in these challenges include studying the pre-processing step to remove possible errors \cite{thulke2023task}, as well as rule-based injection of speech disfluencies in the training data to simulate a spoken interaction \cite{tian2021tod}. While presenting interesting studies, the findings in these studies are limited to the performance of the Automatic Speech Recognition (ASR) models used by the task organizers.

In this work, we investigated the robustness of LLMs to spoken interactions. We used a large dataset of written TODs, MultiWOZ 2.1 \cite{eric2020multiwoz2.1}, to fine-tune T5 and GPT-2 models for the tasks of DST and end-to-end response generation, respectively. We then automatically transcribed the spoken version of a small subset of the dialogues provided by Soltau et al. \cite{soltau2022speech}. We studied and categorized the transcription errors and their distribution by automatically aligning the ASR predictions and the written ground truths. We replicated the same patterns of errors on the large training data based on the observed statistics, as a substitute for a transcribed TOD training set.  Using the obtained noisy training set, we fine-tuned each model for the corresponding task. We investigated the performance of the models when fine-tuned on noisy (ASR-error injected) dialogues, compared to clean (written) dialogues. We evaluated the models on two test sets of Human-Verbatim (human subjects spoke the written turns) and Human-Paraphrased (speakers paraphrased the written turns) TOD dialogues. The main contributions of this work can be summarized as follows:
\begin{itemize}
    \item We transcribed a development set of spoken TODs and studied the error types and distributions observed in ASR transcriptions. We then generated a noisy TOD dataset by injecting these ASR error distributions in a large dataset of written TODs.
    \item We fine-tuned two LLMs for the subtasks of response generation and DST using a dataset of written TODs as well as its noise-injected version;
    \item We evaluated the models' performance on the two subtasks of response generation and DST when fine-tuned on noisy data compared to the clean dialogues. Our analysis includes automatic evaluations via metrics such as model perplexity and Joint Goal Accuracy, as well as human evaluation of the generated responses.
\end{itemize}


\section{Literature Review} 

\textbf{Dialogue State Tracking} There have been various interesting studies on the application of LLMs in DST. Regarding the studies using \textbf{encoder-only} architectures, Feng et al. \cite{feng2022dsgfnet} studied the relations between slots and presented the dialogue state as a graph. The authors used the BERT model as an encoder to embed the dialogue history and the predefined dialogue schema which is then used by a predictor layer to output the dialogue state predictions. Zhang et al. \cite{zhang-etal-2023-monet} studied the task of updating the detected dialogue state throughout the dialogue history by introducing additional noise in the slot values and used a BERT model to encode the dialogue context and state, and a predictor layer based on the similarities of the value and slot vectors in the search space. Regarding the studies that use \textbf{decoder-only} architectures, GPT-2 model has been used by Kulhánek et al. \cite{kulhanek2021augpt}  to perform the task jointly with delexicalized response generation. Furthermore, it has been used by Hosseini-Asl et. al \cite{hosseini2020simpletod} to perform DST jointly with next machine-action prediction as well as delexicalized response generation. Regarding the studies using \textbf{encoder-decoder} architectures, Lee et al. \cite{lee2021dialogue} fine-tuned a T5 model and evaluated its performance in two settings a) generating the whole dialogue state sequentially; and b) generating each slot individually. Lesci et al. \cite{lesci-etal-2023-diable} presented the dialogue state as a table of values and fine-tuned a T5 model to update the table with state values. In another work, Bebensee et al. \cite{bebensee-lee-2023-splat} studied the slot value detection task by selecting a target span for each value and used an encoder-decoder model based on BART with improved attention mechanisms for the task. In a recent work, Won et al. \cite{won2023break} used a combination of LLMs by tasking GPT-2 or T5 with the generation of dialogue state candidates and then using a RoBERTa model for the re-ranking and selecting the correct candidates. While computationally more expensive, this architecture achieved outperforming results compared to other alternatives. The \textbf{datasets} used in the mentioned works are different versions of MultiWOZ 2.0 \cite{budzianowski2018multiwoz}, a collection of multi-domain TODs between crowd workers, with turn-level dialogue state annotations. The different versions of the dataset, such as 2.1 \cite{eric2020multiwoz2.1}, 2.2 \cite{zang2020multiwoz2.2}, 2.3 \cite{han2021multiwoz2.3}, and 2.4 \cite{ye2022multiwoz2.4}, represent attempts to improve the annotation quality such as correcting errors and inconsistencies, the presence of multiple values in the dialogue state, and the introduction of dialogue act tagging. Another TOD dataset used in the literature is Schema Guided Dialogues (SGD) \cite{rastogi2020sdg}, including 16 domains, which is further extended to SGD-X \cite{lee2022sgdx} by adding five paraphrased versions of the domain and slot descriptions. Machines Talking to Machines \cite{shah2018m2m} is another dataset of TODs with two domains.

\textbf{Response Generation} Zhang et al. \cite{zhang-etal-2020-dialogpt} trained a transformer-based architecture identical to GPT-2 for the task of open-domain response generation on a collection of interactions crawled from the Reddit social forum, which can be considered as a distant substitute for dialogue data. The authors showed the model suffers from generating toxic outputs and detecting and controlling such outputs are fundamental. A research trend to address this problem focuses on conditioning the generation on an external piece of information alongside the dialogue history. This piece of information can present an external knowledge piece to the dialogue context in knowledge-grounded generation, or representations of the user model in personalized generations. Regarding the applications of LLMs for \textbf{knowledge-grounded} generation, Hedayatnia et al. \cite{hedayatnia-etal-2020-policy} studied the task of knowledge-grounded response generation using GPT-2 by conditioning the model on a dialogue policy which consists of a selected knowledge pieced, next machine dialogue act, and next response topic. Zhao et al. \cite{zhao-etal-2020-knowledge-grounded} studied knowledge-grounded response generation in open-domain dialogue using the BERT model to select the relevant knowledge piece from Wikipedia for each dialogue, and the GPT-2 model to generate a knowledge-grounded response. Huang et al. \cite{huang2021plato} presented a Transformer-based model for knowledge-grounded response generation with knowledge selection and response generation modules optimized jointly. Regarding \textbf{personalized} response generation,  Wolf et al. \cite{wolftransfertransfo} and Kasahara et al. \cite{Kasahara} fine-tuned the GPT-2 model for the task of response generation grounded on persona statements. In a more recent work, Mousavi et al. \cite{mousavi-etal-2023-response} studied the task of response generation in longitudinal dialogues, conditioning the model on the extracted personal knowledge model of the user.

\textbf{Robustness for Spoken Dialogues} To address the lack of a proper spoken dialogue dataset for DST and response generation, Kim et al. \cite{kim2021robust} collected and transcribed 107 spoken human-human conversations following the MultiWOZ schema. This data was then used in DSTC10 as a test set to evaluate dialogue systems that are trained on written dialogues on spoken conversations. Thulke et al. \cite{thulke2023task} used this data to study the impact of pre-processing steps to increase the robustness of the dialogue system to speech recognition noise and obtain better scores. In another work, this benchmark was used by Tian et al. \cite{tian2021tod}, where the authors developed a rule-based algorithm to inject ASR errors in the written training data to improve performance and participate in the DSTC10 challenge. Nevertheless, the test set consisted only of the transcribed version of the dialogues which limits the error studies and findings to the performance of the ASR model used to transcribe the dialogues. In another work, Soltau et al. \cite{soltau2022speech} collected a spoken version of MultiWOZ 2.1 for the task of spoken DST and response generation. The authors shared with the community the speech audio files and transcriptions in 3 spoken scenarios: a) using a Text-to-Speech model (for train, valid, and test sets); b) a Human-Verbatim version using crowd-workers (for valid and test sets); and c) a Human-Paraphrased version using crowd-workers (for test set only). This data was then used as a benchmark in DSTC 11. While the aforementioned works introduced small-scale spoken TODs, in a most recent work Si et al. \cite{si2023spokenwoz} collected and annotated a spoken human-human dialogue dataset consisting of 8 domains and 5.7k dialogues among 250 different participants. Furthermore, while not focusing on dialogue, in another attempt to address the gap between written and spoken language Wang et al. \cite{wang2023slm} developed a Speech and Language model based on a speech encoder and a pre-trained LLM and evaluated its performance in speech recognition and speech translation tasks.


\section{Approach}

We studied whether LLMs are robust for spoken interactions, and whether fine-tuning the LLMs on ASR-noise-injected dialogues can improve their performance for such dialogues. We transcribed a small number of spoken TODs and studied the transcription errors by automatically aligning the predictions and the ground truths. The observed error pattern was then injected into a large dataset of TODs as a training set with simulated noise.

\subsection{Dataset}

The dataset used in this work is MultiWOZ 2.1 \cite{eric2020multiwoz2.1}, a large dataset of written multi-domain TODs. This dataset consists of 10k dialogues (8k, 1k, and 1k dialogues for train, dev, and test splits) with an average number of 14 turns per dialogue, spanning over 6 domains. The data includes annotations about the dialogue state and the dialogue acts at the turn level. In addition to the MultiWOZ 2.1 original splits, we used the spoken version of the development sets provided by Soltau et al \cite{soltau2022speech} in two different settings as a) Human-Verbatim (HV): where the crowd-workers were asked to speak the written user turns in the dev and test splits; and b) Human-Paraphrased (HP): where the crowd-workers were asked to speak a paraphrased version of the written user turns in the test set dialogues. Thus, the data used in our experiments are 1) written training set; 2) development set as written and HV; and 3) test set as written, HV, and HP. Note that the slot values in the spoken development sets are different from the original MultiWOZ 2.1 evaluation sets as Soltau et al. \cite{soltau2022speech} replaced the slot values with new values unseen in the training set. The authors reported that the replacements were done at the dialogue level and included location names, hotel and restaurant names, and well as time slots.

\subsection{Sub-Tasks \& Models}

\textbf{Dialogue State Tracking} We studied the robustness of LLM for spoken DST, where the model is tasked to keep track of the slot values necessary for fulfilling the user's intents. The values can be inferred from the dialogue history, extracted from the user responses, or proposed by the system and confirmed by the user. The LLM fine-tuned for this task is T5 Small \cite{limitsoftransfer}, (12 layers, 60M parameters), a transformer-based encoder-decoder model, pre-trained on the Common Crawl dataset with 750GB of web page text. The fine-tuning was performed following the approach proposed by Lee et al. \cite{lee2021dialogue} to generate a value for each slot individually.

\textbf{Response Generation} In the second subtask we evaluated the robustness of LLMs in end-to-end response generation for spoken TODs. In this subtask, the model is required to encode the dialogue history and generate an appropriate and coherent response with respect to the dialogue context, to assist the user in fulfilling her intentions. The LLMs used for this task are GPT-2 small (12 layers of decoder blocks, 117M parameters), and GPT-2 medium (24 layers of decoder blocks, 345M parameters) \cite{radford2019gpt2}, as decoder-only models pre-trained on WebText dataset with a 40GB web crawled documents.

\subsection{ASR-Error Simulation}

To obtain a dataset of spoken TODs for fine-tuning the models, we studied the speech errors observed in the transcriptions of the Human-Verbatim (HV) development set as a small-scale dataset of spoken dialogues. We developed a methodology to replicate the same pattern in the training set. Firstly, we transcribed the dialogues via the SpeechBrain toolkit \cite{speechbrain} using Whisper tiny model \cite{radford2023whisper}, a transformer-based encoder-decoder (4 layers, 39M parameters) pre-trained on 680k hours of speech data. The transcription was followed by automatic alignment of the predictions and the ground truths, via NIST scoring toolkit\footnote{SCTK, the NIST Scoring Toolkit: \url{https://github.com/usnistgov/SCTK}}, by minimising the Levenshtein distance.

Using NIST toolkit, we defined three main categories for the errors: a) Insertions: one or more additional words can be found in the transcription, compared to ground truth; b) Deletions: one or more words are missing from the transcription, compared to ground truth; c) Substitutions: one or more words in the transcription are different from what is reported in the ground truth. Table \ref{tab:whisper-wer-results} presents the percentage of each error category observed in the transcriptions of the HV development set, as well as the observed Word Error Rate (WER), and Sentence Error Rate (SER). The WER is computed by summing all the errors in the transcription, over the total number of words. The SER is computed by the number of sentences containing at least one error over the total number of sentences.  

\begin{table}[t]
\centering
    \caption{The percentage of error categories observed in the transcriptions of the Human-Verbatim development set, compared to the percentage of injected errors by our algorithm, observed a posteriori.}
    \label{tab:whisper-wer-results}   
    \begin{tabular}{p{3cm}p{1.5cm}p{1.5cm}p{2.0cm}p{1.0cm}p{1.0cm}}
    \textbf{Data} & \textbf{Insertions} & \textbf{Deletions} & \textbf{Substitutions} & \textbf{WER} & \textbf{SER} \\
        \hline\hline\noalign{\smallskip}
        Human-Verbatim & \multirow{2}{*}{2.3} & \multirow{2}{*}{1.9} & \multirow{2}{*}{8.1} & \multirow{2}{*}{12.3} & \multirow{2}{*}{54.9} \\
        Transcriptions \\
        \hline\noalign{\smallskip}
         MultiWOZ 2.1\\
        \hspace{0.25cm}{\textit{Error-Injected Train}} & 2.1 & 5.9 & 8.0 & 16.1 & 89.2 \\
        \hspace{0.25cm}{\textit{Error-Injected Dev}}   & 2.1 & 6.0 & 8.3 & 16.4 & 90.3 \\
    \end{tabular}
\end{table}

To obtain the training data for spoken TODs, we calculated the statistics of errorful recognition for each token in the transcriptions, as well as the distribution of each error type for the specific token. For instance, the token ``\textit{book}" has been correctly recognized in 94.6\% of the cases in the transcriptions. Nevertheless, in the remaining cases, it was confused with ``\textit{look}" (2.2\%) and ``\textit{put}" (2\%). Meanwhile, the word ``\textit{post-code}" is correctly recognized in 68.9\%, while it is confused with ``\textit{code}" (22\%) and ``\textit{post-card}" (3\%) in the remaining cases. We then developed an algorithm to automatically inject the observed ASR error in the MultiWOZ training set. For each token in the training data, the algorithm counts the number of occurrences of the token in the data and alternates the token with the corresponding ASR error with the same ratio and distribution calculated beforehand. As a point of reference, we also injected errors in the MultiWOZ development set to compare with the obtained ASR transcription of the same dialogues. 

The results of the ASR-error injection on the MultiWOZ dialogues, presented in \ref{tab:whisper-wer-results}, indicate the percentage of Insertions and Substitution applied by the injection algorithm is approximately equal to the percentage of errors observed in the transcriptions. Nevertheless, the percentage of Deletions, and consequently the WER and SER, is considerably higher in the error-injected data sets. By conducting further analysis, we noticed the percentage of injected errors is in line with the distributions calculated a priori. However, the Whisper pipeline includes punctuation removal, tokenization, and sequence normalization. Therefore, various differences between the MultiWOZ sequence and the normalized version are associated with the Deletion error category. For instance, while 72\% of the sentences in the user turns contain punctuation, to comply with the observed patterns in the transcription, 66\% of the punctuations are removed after the noise-injection step. 

\section{Evaluation}
We evaluated the robustness of the models and compared their performance when fine-tuned on clean and noisy dialogues in two subtasks of DST and response generation.

\subsection{Dialogue State Tracking}

In this task, we fine-tuned T5 small and evaluated its performance using Joint Goal Accuracy (JGA). JGA measures the percentage of turns in the dataset in which the predicted dialogue state perfectly matches the ground truth. The fine-tuning was performed using AdamW optimizer \cite{loshchilov2018decoupled} with a learning rate of $5\cdot 10^{-5}$, for a maximum of 10 epochs and with an early-stopping patience counter of 3 epochs. 

By fine-tuning on the clean (MultiWOZ) TODs, we reached the JGA of 53.61, which is in line with the performance observed in the literature by Lee et al. \cite{lee2021dialogue}. However, we observed a performance drop from 53.61 to 21.19 JGA points by the outperforming model on the spoken test sets HV and HP. This is due to the fact that the slot values in the spoken development sets are different from the original MultiWOZ 2.1 evaluation sets (this drop in performance is in line with what Soltau et al. \cite{soltau2022speech} reported as the state-of-art for the new development sets). Therefore, we consider this performance as the baseline to investigate whether introducing speech error in the training data results in improvements.

The performance of the models on the two spoken test sets, presented in Table \ref{tab:t5_jga_postprocess}, indicates that introducing speech noise in the training data slightly worsens the model performance on spoken dialogues. Besides, the results suggest that paraphrased user requests (HP) are slightly more challenging for the model compared to the HV.

Since the task of DST is to extract and select the slot values throughout the dialogue with the user, in our next analysis we decided to introduce additional noise specifically on the slot values. For this purpose, we selected 20\% and 50\% of the slot values in the noisy dialogues and replaced them with their errorful recognition observed in the transcriptions \footnote{The slots with additional noise include \textit{attraction-name, hotel-name, restaurant-name, taxi-departure, taxi-destination, train-departure, train-destination}.}. The results suggest that introducing noise on slot values specifically can result in an increase in the model robustness to the observed noise in the spoken dialogues. Nevertheless, the outperforming model remains to be the one trained on clean dialogues.

\begin{table}[t]
\centering
    \caption{The performance (Joint Goal Accuracy) of the T5 model for DST, indicating that introducing noise in the training data slightly worsens the model's performance. Nevertheless, introducing noise on the slot values increases the model robustness to noisy data slightly.}
    \label{tab:t5_jga_postprocess}      
    \begin{tabular}{p{5cm}p{1.6cm}p{1.7cm}}
    
        \multirow{1}{*}{\textbf{Models}}& \multicolumn{1}{c}{\textbf{Human-Verb.}} & \multicolumn{1}{c}{\textbf{Human-Par.}}  \\
        \hline\hline\noalign{\smallskip}
        T5 Small Fine-Tuned on &&\\
        \hspace{0.25cm}{\textit{Clean Dialogues}} & \textbf{21.19 }& \textbf{19.93} \\
        \hspace{0.25cm}{\textit{Noisy Dialogues}} & 19.07 & 18.08 \\
        \hspace{0.25cm}{\textit{Noisy Dialogues + 20\% of Slot Values}}& 19.72 & 18.38 \\
        \hspace{0.25cm}{\textit{Noisy Dialogues + 50\% of Slot Values}} & 20.09 & 18.73 \\
     \hline\hline\noalign{\smallskip}
    \end{tabular}
\end{table}

\subsection{Response Generation}

For response generation, we fine-tuned GPT-2 Small and Medium\footnote{We did not experiment with the Large model due to computation resource limitations.} using AdamW optimizer \cite{loshchilov2018decoupled}, with a linearly decaying learning rate scheduler, with early-stopping patience of 3 epochs, and batch size of $16$. We performed hyper-parameter optimization using the development sets to choose the best values for the learning rate, history window, gradient clipping, and number of gradient accumulation steps. Fine-tuning on clean TODs, the best performance on the unseen test set of clean dialogues was $ppl=3.94$ for GPT-2 Small and $ppl=3.85$ for GPT-2 medium.

\begin{figure*}[!t]
    \centering
    \includegraphics[width=\textwidth]{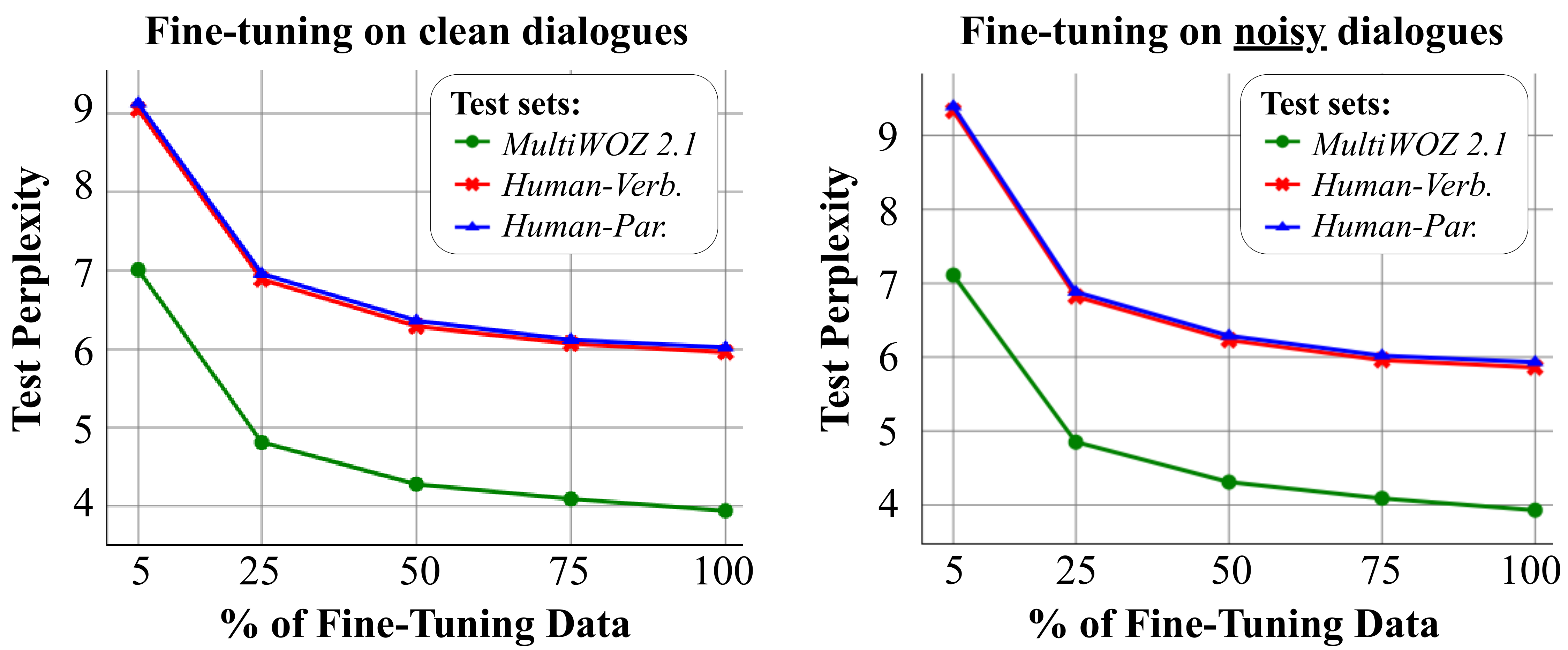}
    \caption{The influence of fine-tuning the model (GPT-2 Small) using different portions of the clean and noisy dialogues. The model is evaluated on three test sets of clean (MultiWOZ 2.1) dialogues, HV, and HP.}
    \label{fig:FTplots}
\end{figure*}

\subsubsection{Automatic Evaluation}

Given the relatively small size of the fine-tuning data compared to the pre-training data, we investigated the effect of fine-tuning the models by incrementally augmenting the fine-tuning set. Figure \ref{fig:FTplots} presents the perplexity (\textit{ppl}) trend of the GPT-2 Small model when fine-tuned on different portions of the clean and noisy dialogues on the three unseen test sets. This augmentation of the training dataset was carried out progressively, ensuring that smaller segments were contained within the larger ones. As shown in the plots, the model performance improves considerably upon observing the initial 25\% of the training set, indicating that the fine-tuning process was particularly effective. Afterward, the model demonstrated a steady improvement by observing the rest of the data. Besides, the perplexity trends in Figure \ref{fig:FTplots} indicate that regardless of the fine-tuning data, the model has a similar perplexity on the spoken test sets, which is comparably higher than the perplexity on clean (MultiWOZ) dialogues. This suggests that generating a response for the clean dialogues is less challenging.

\begin{table}[!t]
\centering
    \caption{Automatic evaluation of fine-tuned GPT-2 models averaged over 5 runs for each setting. The results indicate that introducing speech noise in the fine-tuning data slightly improves the model performance in terms of \textit{ppl}.}
    \label{tab:response-gen-auto-eval-ppl}    
    \begin{tabular}{p{3.5cm}p{1.2cm}p{1.15cm}|p{1.4cm}p{1.4cm}}
         \multirow{2}{*}{\textbf{Models}}& \multicolumn{2}{c}{\textbf{Human-Verb.}} & \multicolumn{2}{c}{\textbf{Human-Par.}}  \\
         & \textit{\textbf{nll}} & \textit{\textbf{ppl}}& \textit{\textbf{nll}} & \textit{\textbf{ppl}}  \\
        \hline\hline\noalign{\smallskip}
        GPT-2 Small &&&&\\
        \hspace{0.5cm}{\textit{Clean Fine-Tuning}} & 1.79 & 5.96 & 1.79 & 6.02 \\
        \hspace{0.5cm}{\textit{Noisy Fine-Tuning}} & 1.77 & 5.86 & 1.78 & 5.93 \\
         \hline \noalign{\smallskip}
         GPT-2 Medium &&&&\\
        \hspace{0.5cm}{\textit{Clean Fine-Tuning}} & 1.72 & 5.60 & 1.73 & 5.65 \\
         \hspace{0.5cm}{\textit{Noisy Fine-Tuning}}& \textbf{1.71} & \textbf{5.55} & \textbf{1.72} & \textbf{5.60} \\
         \hline\hline
    \end{tabular}
\end{table}

The automatic performance evaluations on the two test sets, presented in Table \ref{tab:response-gen-auto-eval-ppl}, suggest that fine-tuning the models on the noisy dialogues can slightly improve the models' perplexity. Besides, the results show that the model perplexity is higher by a negligible amount on the paraphrased test set, compared to HV. Nevertheless, the Medium model outperforms the Small alternative in all settings. Therefore, we select the Medium model and focus the rest of our analysis on the performance of this model size.  

\begin{figure*}[t]
    \centering
    \includegraphics[width=0.9\textwidth]{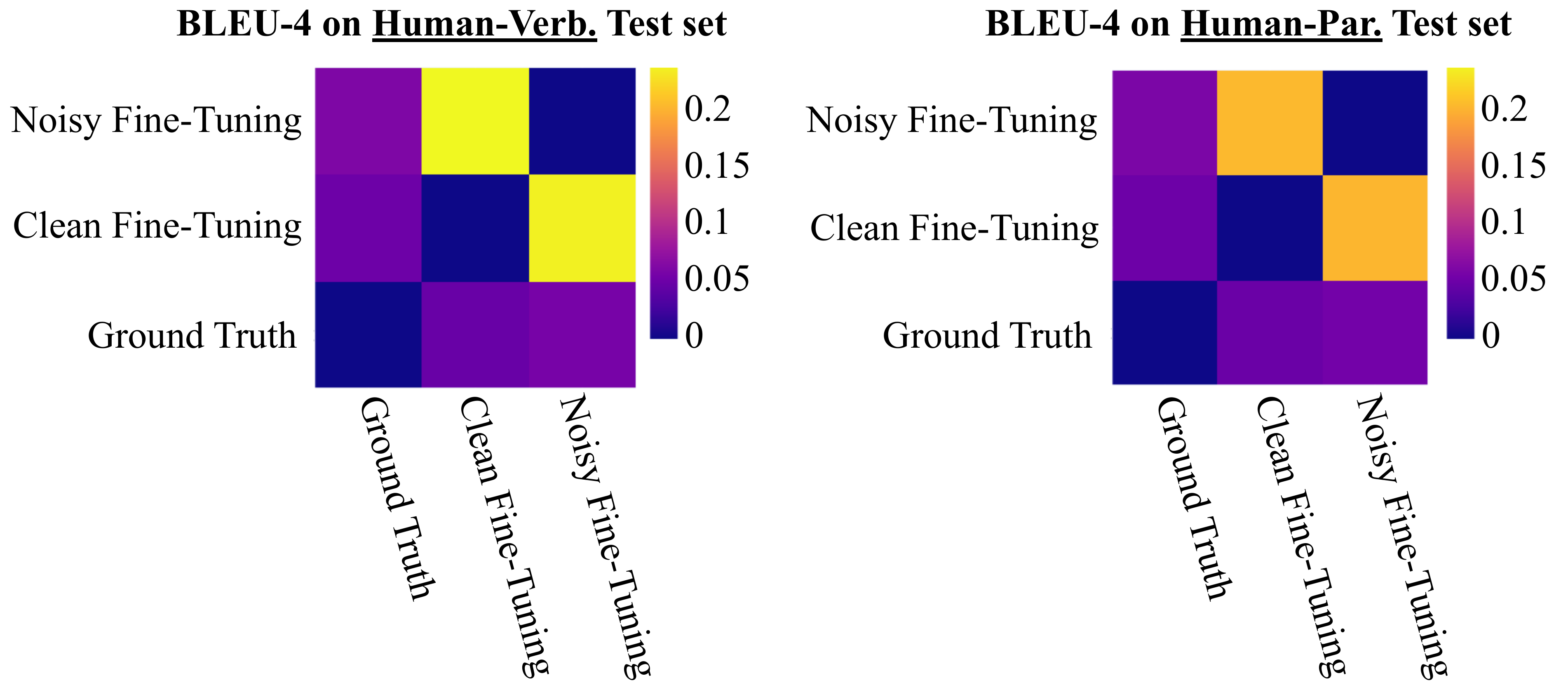}
    \caption{The lexical similarity among the generated responses (GPT-2 Medium) and ground truth in different fine-tuning settings. The lowest lexical similarity is among generated responses and the ground truth.}
    \label{fig:B4plots}
\end{figure*}

We evaluated the lexical similarity among the generated responses by GPT-2 Medium and the ground truth in different fine-tuning settings using BLEU-4. The results presented in Figure \ref{fig:B4plots}, show that regardless of the test set, the lowest lexical similarity is among the generated responses and the ground truths.

\subsubsection{Human Evaluation}
We sampled 20 dialogues from each test sets, HV and HP, to conduct human evaluations. For each sampled dialogue, we evaluated the generated responses by the GPT-2 medium fine-tuned on clean and noisy dialogues as well as the ground truth, as a point of reference. Using the human evaluation protocol proposed by \cite{humeval}, the annotators were asked to assess each response with respect to three criteria as \textit{Correctness}, \textit{Appropriateness}, and \textit{Contextualization} using three options (positive, negative, or ``I don’t know"). For the negative decisions ``\textit{Not Appropriate}" and ``\textit{Not Contextualization}", the annotators were further asked to explain their decisions by highlighting any potential errors in the generated response. Each candidate was reviewed by 5 crowd-workers\footnote{42 crowd-workers were recruited through Prolific crowd-sourcing platform, of which 2 workers were not qualified.} and the overall Inter Annotator Agreement (IAA) measured using Fleiss’ $\kappa$ was 0.24 (=Fair agreement).  

\begin{table*}[t]
\centering
\small
\caption{Human Evaluation of the models fine-tuned on clean and noisy TODs on both testsets of Human-Verbatim (H-Verb.) and Human-Paraphrased (H-Par.). The responses generated by the model fine-tuned on noisy TODs are more appropriate and contextualized, compared to the model fine-tuned on clean dialogues.}
\begin{tabular}{p{2.9cm}|p{1.25cm}p{1.25cm}|p{1.25cm}p{1.25cm}|p{1.25cm}p{1.25cm}}

\multirow{3}{*}{\textbf{Models}}  & \multicolumn{6}{c}{\textbf{Human Evaluation}} \\
  & \multicolumn{2}{c}{\textbf{Correctness}} & \multicolumn{2}{c}{\textbf{Appropriateness}} & \multicolumn{2}{c}{\textbf{Contextualization}} \\
  \cline{2-7}\noalign{\smallskip}
  & H-Verb.&H-Par.& H-Verb.&H-Par.& H-Verb.&H-Par.\\
          \hline\hline\noalign{\smallskip}
         Ground Truth &0.85& 0.70&1.00&1.00&1.00&0.95 \\
          \hline\noalign{\smallskip}
GPT-2 Medium &&&&&&\\
       \hspace{0.5cm}{\textit{Clean Fine-Tuning}} &\textbf{1.00}&0.85&0.95&0.80&0.90&0.85 \\
         \hspace{0.5cm}{\textit{Noisy Fine-Tuning}} &0.95&0.85&0.95&\textbf{0.95}&\textbf{0.95}&\textbf{0.95} \\
\hline \hline
\end{tabular}
\label{table:he}
\end{table*}

\begin{figure*}[!t]
    \centering
    \includegraphics[width=\textwidth]{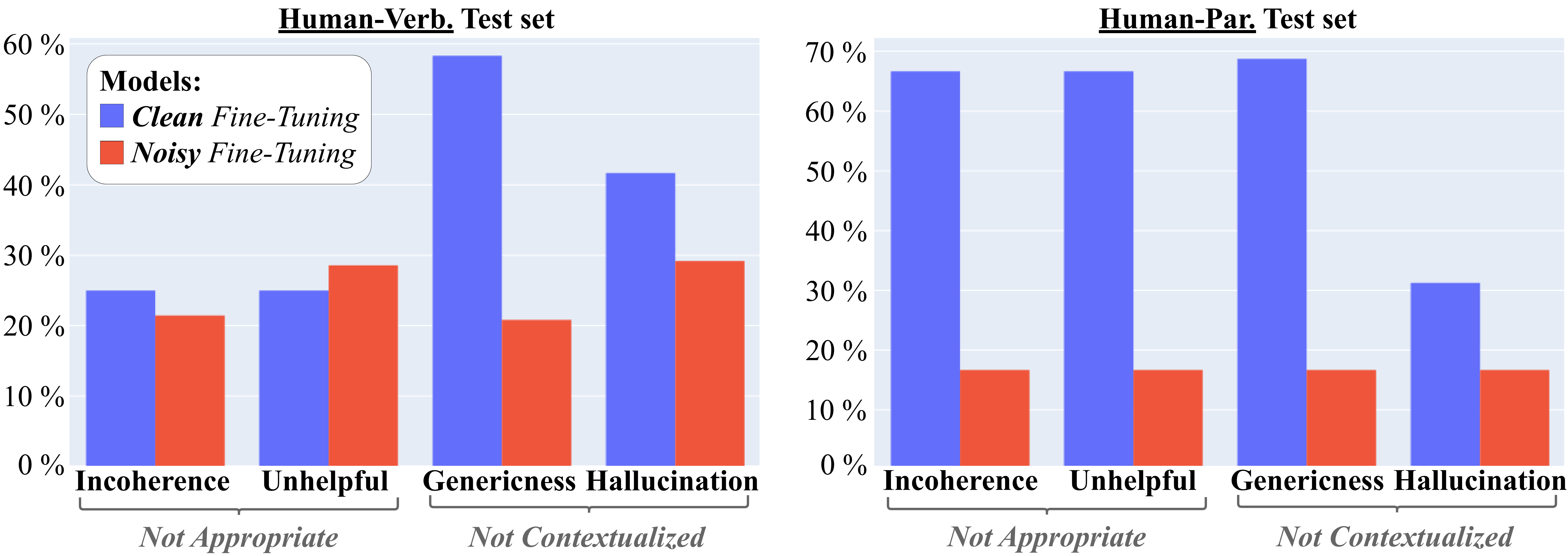}
    \caption{The error types selected by the human annotators to explain their negative judgments (\textit{Not Appropriate}, and \textit{Not Contextualized}) of the response candidates.}
    \label{fig:exp}
\end{figure*}

The results of this evaluation are presented in Table \ref{table:he}. Regarding the \textit{\textbf{Correctness}} of the candidates, the results show that the model fine-tuned on clean dialogues makes no mistakes on the verbatim test set, while the outputs of both models contain grammatical and structural errors on the paraphrased test set. Furthermore, the results suggest that there are structural and grammatical errors in the ground truth turns (test sets are crowd-sourced) and the models outperform the ground truth in this criterion only. With respect to \textit{\textbf{Appropriateness}} criterion, both models have a similar performance on HV test set, the model fine-tuned on noisy data produces more appropriate responses by 15\%. Similarly, for the \textit{\textbf{Contextualization}} criterion, the results indicate that fine-tuning on noisy data helps the model to produce more contextualized responses for both verbatim and paraphrased spoken settings.

In the next analysis, we studied the explanations the crowd-workers provided to motivate their judgments following the same categorization proposed by Mousavi et al. \cite{humeval}. The frequency of the error types detected by the judges in the generated responses is presented in Figure \ref{fig:exp}. A common trend observed in both test sets is that the model fine-tuned on noisy dialogues generates less generic, less hallucinated, and more coherent responses.  Furthermore, noisy fine-tuning helps the model generate fewer responses that are unhelpful for TODs on the paraphrased test set. 

\begin{figure*}[!t]
    \centering
    \includegraphics[width=\textwidth]{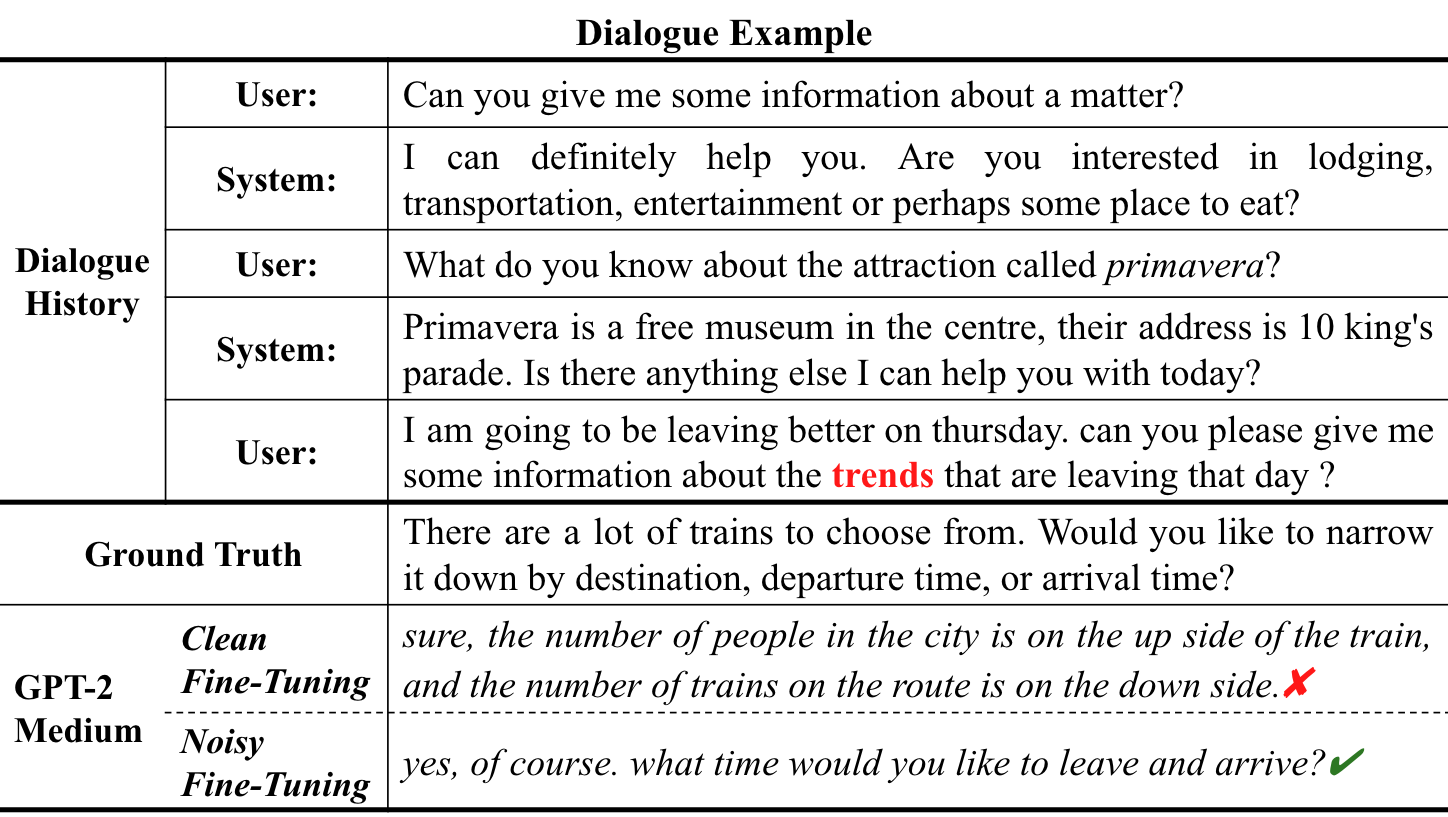}
    \caption{A dialogue example from the Human-Paraphrased test set with wrong recognition of the word "trains" (wrongly recognized as "trends" by the ASR model). While the model fine-tuned on clean dialogues fails to handle this error, the model fine-tuned on noisy TODs is more robust to such errors and generates an appropriate response.}
    \label{fig:exam}
\end{figure*}

Figure \ref{fig:exam} shows an example of the dialogues in the HP test set. The word "\textit{trains}" in the last user turn in the dialogue history is mistakenly recognized as "\textit{trends}" by the ASR model. Due to this error in recognition, the model fine-tuned on clean dialogues fails to generate an appropriate response. However, the model fine-tuned on noisy dialogues shows a higher level of robustness and handles this noise, resulting in an appropriate and contextualized response generation.

\section{Conclusion}

In this work, we studied the robustness of LLMs for spoken TODs by fine-tuning two LLMs, GPT-2 and T5, on clean and noise-injected task-oriented dialogues. We evaluated the performance of the fine-tuned models using two spoken test sets of Human-Verbatim and Human-Paraphrased in two tasks of end-to-end response generation and dialogue state tracking. 

Our evaluation showed that a) Human-paraphrased dialogues are slightly more challenging as the performance of the models is lower compared to the Human-Verbatim test set in both tasks. This difference in the performance can be due to the fact that paraphrasing is a cognitively loaded task and therefore the obtained test set includes more spontaneous events such as disfluency, etc; b) Fine-tuning on noisy dialogues does not increase the robustness of the model in dialogue state tracking (according to Joint Goal Accuracy). Nevertheless, introducing noise specifically on the slot values showed a slight increase in performance; and c) Regarding the end-to-end response generation task, noisy fine-tuning of the model results in a negligible improvement in perplexity score and considerable improvement judged by human annotators. More specifically, the model fine-tuned on noisy TODs generates less hallucinated or generic responses and obtained higher scores for appropriateness and contextualization on both test sets. 
Therefore, while LLMs are not robust to spoken noise by default, our evaluations suggest that fine-tuning/training such models on a proper dataset of spoken TODs can result in a more robust performance. Besides, outcome of the human evaluation of the responses once again indicates the limitations and uninterpretability of automatic metrics (such as \textit{ppl}) and their low correlation with human judgments.

\section*{Acknowledgements}
We acknowledge the support of the MUR PNRR project FAIR - Future AI Research (PE00000013) funded by the NextGenerationEU. We acknowledge the support of the Digital Research Alliance of Canada (alliancecan.ca) for contributing to some of the experiments in this research. We would like to thank Prof. Renato De Mori (McGill University) for the fruitful discussions.

%
%

\begin{thebibliography}{99.}%

\bibitem{bebensee-lee-2023-splat} Bebensee, B., \& Lee, H. (2023). Span-Selective Linear Attention Transformers for Effective and Robust Schema-Guided Dialogue State Tracking. In Proceedings of the 61st Annual Meeting of the Association for Computational Linguistics (Volume 1: Long Papers) (pp. 78–91). Association for Computational Linguistics.

\bibitem{budzianowski2018multiwoz} Budzianowski, P., Wen, T. H., Tseng, B. H., Casanueva, I., Ultes, S., Ramadan, O., \& Gasic, M. (2018). MultiWOZ-A Large-Scale Multi-Domain Wizard-of-Oz Dataset for Task-Oriented Dialogue Modelling. In Proceedings of the 2018 Conference on Empirical Methods in Natural Language Processing (pp. 5016-5026).

\bibitem{eric2020multiwoz2.1} Eric, M., Goel, R., Paul, S., Sethi, A., Agarwal, S., Gao, S., ... \& Hakkani-Tur, D. (2020, May). MultiWOZ 2.1: A Consolidated Multi-Domain Dialogue Dataset with State Corrections and State Tracking Baselines. In Proceedings of the Twelfth Language Resources and Evaluation Conference (pp. 422-428).

\bibitem{feng2022dsgfnet} Feng, Y., Lipani, A., Ye, F., Zhang, Q., \& Yilmaz, E. (2022, May). Dynamic Schema Graph Fusion Network for Multi-Domain Dialogue State Tracking. In Proceedings of the 60th Annual Meeting of the Association for Computational Linguistics (Volume 1: Long Papers) (pp. 115-126).


\bibitem{han2021multiwoz2.3} Han, T., Liu, X., Takanabu, R., Lian, Y., Huang, C., Wan, D., ... \& Huang, M. (2021). Multiwoz 2.3: A multi-domain task-oriented dialogue dataset enhanced with annotation corrections and co-reference annotation. In Natural Language Processing and Chinese Computing: 10th CCF International Conference, NLPCC 2021, Qingdao, China, October 13–17, 2021, Proceedings, Part II 10 (pp. 206-218). Springer International Publishing.

\bibitem{hedayatnia-etal-2020-policy} Hedayatnia, B., Gopalakrishnan, K., Kim, S., Liu, Y., Eric, M., \& Hakkani-Tur, D. (2020, December). Policy-Driven Neural Response Generation for Knowledge-Grounded Dialog Systems. In Proceedings of the 13th International Conference on Natural Language Generation (pp. 412-421).


\bibitem{hosseini2020simpletod} Hosseini-Asl, E., McCann, B., Wu, C. S., Yavuz, S., \& Socher, R. (2020). A simple language model for task-oriented dialogue. Advances in Neural Information Processing Systems, 33, 20179-20191.

\bibitem{huang2021plato} Huang, X., He, H., Bao, S., Wang, F., Wu, H., \& Wang, H. (2021, November). Plato-kag: Unsupervised knowledge-grounded conversation via joint modeling. In Proceedings of the 3rd Workshop on Natural Language Processing for Conversational AI (pp. 143-154).


\bibitem{Kasahara} Kasahara, T., Kawahara, D., Tung, N., Li, S., Shinzato, K., \& Sato, T. (2022, July). Building a Personalized Dialogue System with Prompt-Tuning. In Proceedings of the 2022 Conference of the North American Chapter of the Association for Computational Linguistics: Human Language Technologies: Student Research Workshop (pp. 96-105).

\bibitem{bert} Kenton, J. D. M. W. C., \& Toutanova, L. K. (2019). BERT: Pre-training of Deep Bidirectional Transformers for Language Understanding. In Proceedings of NAACL-HLT (pp. 4171-4186).

\bibitem{kim2021robust} Kim, S., Liu, Y., Jin, D., Papangelis, A., Gopalakrishnan, K., Hedayatnia, B., \& Hakkani-Tür, D. (2021, December). “How Robust R U?”: Evaluating Task-Oriented Dialogue Systems on Spoken Conversations. In 2021 IEEE Automatic Speech Recognition and Understanding Workshop (ASRU) (pp. 1147-1154). IEEE.

\bibitem{kulhanek2021augpt} Kulhánek, J., Hudeček, V., Nekvinda, T., \& Dušek, O. (2021, November). AuGPT: Auxiliary Tasks and Data Augmentation for End-To-End Dialogue with Pre-Trained Language Models. In Proceedings of the 3rd Workshop on Natural Language Processing for Conversational AI (pp. 198-210).

\bibitem{lee2021dialogue} Lee, C. H., Cheng, H., \& Ostendorf, M. (2021, November). Dialogue State Tracking with a Language Model using Schema-Driven Prompting. In Proceedings of the 2021 Conference on Empirical Methods in Natural Language Processing (pp. 4937-4949).

\bibitem{lee2022sgdx} Lee, H., Gupta, R., Rastogi, A., Cao, Y., Zhang, B., \& Wu, Y. (2022, June). SGD-X: A Benchmark for Robust Generalization in Schema-Guided Dialogue Systems. In Proceedings of the AAAI Conference on Artificial Intelligence (Vol. 36, No. 10, pp. 10938-10946).

\bibitem{lesci-etal-2023-diable} Lesci, P., Fujinuma, Y., Hardalov, M., Shang, C., Benajiba, Y., \& Marquez, L. (2023). Diable: Efficient Dialogue State Tracking as Operations on Tables. In Findings of the Association for Computational Linguistics: ACL 2023 (pp. 9697–9719). Association for Computational Linguistics.


\bibitem{loshchilov2018decoupled} Loshchilov, I., \& Hutter, F. (2018, September). Decoupled Weight Decay Regularization. In International Conference on Learning Representations.


\bibitem{humeval} Mousavi, S. M., Roccabruna, G., Lorandi, M., Caldarella, S., \& Riccardi, G. (2022, December). Evaluation of Response Generation Models: Shouldn’t It Be Shareable and Replicable?. In Proceedings of the 2nd Workshop on Natural Language Generation, Evaluation, and Metrics (GEM) (pp. 136-147).

\bibitem{mousavi-etal-2023-response} Mousavi, S., Caldarella, S., \& Riccardi, G. (2023). Response Generation in Longitudinal Dialogues: Which Knowledge Representation Helps?. In Proceedings of the 5th Workshop on NLP for Conversational AI (NLP4ConvAI 2023) (pp. 1–11). Association for Computational Linguistics.

\bibitem{radford2019gpt2} Radford, A., Wu, J., Child, R., Luan, D., Amodei, D., \& Sutskever, I. (2019). Language models are unsupervised multitask learners. OpenAI blog, 1(8), 9.

\bibitem{radford2023whisper} Radford, A., Kim, J. W., Xu, T., Brockman, G., McLeavey, C., \& Sutskever, I. (2023, July). Robust speech recognition via large-scale weak supervision. In International Conference on Machine Learning (pp. 28492-28518). PMLR.


\bibitem{limitsoftransfer} Raffel, C., Shazeer, N., Roberts, A., Lee, K., Narang, S., Matena, M., ... \& Liu, P. J. (2020). Exploring the limits of transfer learning with a unified text-to-text transformer. The Journal of Machine Learning Research, 21(1), 5485-5551.

\bibitem{rastogi2020sdg} Rastogi, A., Zang, X., Sunkara, S., Gupta, R., \& Khaitan, P. (2020, April). Towards scalable multi-domain conversational agents: The schema-guided dialogue dataset. In Proceedings of the AAAI conference on artificial intelligence (Vol. 34, No. 05, pp. 8689-8696).

\bibitem{speechbrain} Ravanelli, M., Parcollet, T., Plantinga, P., Rouhe, A., Cornell, S., Lugosch, L., ... \& Bengio, Y. (2021). SpeechBrain: A general-purpose speech toolkit. arXiv preprint arXiv:2106.04624.


\bibitem{shah2018m2m} Shah, P., Hakkani-Tür, D., Tür, G., Rastogi, A., Bapna, A., Nayak, N., \& Heck, L. (2018). Building a conversational agent overnight with dialogue self-play. arXiv preprint arXiv:1801.04871. 

\bibitem{si2023spokenwoz} Si, S., Ma, W., Wu, Y., Dai, Y., Gao, H., Lin, T. E., ... \& Li, Y. (2023). SpokenWOZ: A Large-Scale Speech-Text Benchmark for Spoken Task-Oriented Dialogue in Multiple Domains. arXiv preprint arXiv:2305.13040.

\bibitem{soltau2022speech} Soltau, H., Shafran, I., Wang, M., Rastogi, A., Zhao, J., Jia, Y., ... \& Miranda, A. (2022). Speech Aware Dialog System Technology Challenge (DSTC11). arXiv preprint arXiv:2212.08704.

\bibitem{tian2021tod} Tian, X., Huang, X., He, D., Lin, Y., Bao, S., He, H., ... \& Wang, H. (2021). TOD-DA: towards boosting the robustness of task-oriented dialogue modeling on spoken conversations. arXiv preprint arXiv:2112.12441.

\bibitem{thulke2023task} Thulke, D., Daheim, N., Dugast, C., \& Ney, H. (2023). Task-Oriented Document-Grounded Dialog Systems by HLTPR@ RWTH for DSTC9 and DSTC10. IEEE/ACM Transactions on Audio, Speech, and Language Processing.


\bibitem{attentionpaper} Vaswani, A., Shazeer, N., Parmar, N., Uszkoreit, J., Jones, L., Gomez, A. N., ... \& Polosukhin, I. (2017). Attention is all you need. Advances in neural information processing systems, 30.

\bibitem{wang2023slm} Wang, M., Han, W., Shafran, I., Wu, Z., Chiu, C. C., Cao, Y., ... \& Wu, Y. (2023). SLM: Bridge the thin gap between speech and text foundation models. arXiv preprint arXiv:2310.00230.


\bibitem{wolftransfertransfo} Wolf, T., Sanh, V., Chaumond, J., \& Delangue, C. (2019). Transfertransfo: A transfer learning approach for neural network based conversational agents. arXiv preprint arXiv:1901.08149.

\bibitem{won2023break} Won, S., Kwak, H., Shin, J., Han, J., \& Jung, K. (2023, July). BREAK: Breaking the Dialogue State Tracking Barrier with Beam Search \& Re-ranking. In Proceedings of the 61st Annual Meeting of the Association for Computational Linguistics (Volume 1: Long Papers) (pp. 2832-2846).

\bibitem{ye2022multiwoz2.4} Ye, F., Manotumruksa, J., \& Yilmaz, E. (2022, September). MultiWOZ 2.4: A Multi-Domain Task-Oriented Dialogue Dataset with Essential Annotation Corrections to Improve State Tracking Evaluation. In Proceedings of the 23rd Annual Meeting of the Special Interest Group on Discourse and Dialogue (pp. 351-360).

\bibitem{e2eres} Yu, X., Wu, Q., Qian, K., \& Yu, Z. (2022). Reinforced Language Modeling for End-to-End Task Oriented Dialog. arXiv preprint arXiv:2211.16773.

\bibitem{zang2020multiwoz2.2} Zang, X., Rastogi, A., Sunkara, S., Gupta, R., Zhang, J., \& Chen, J. (2020, July). MultiWOZ 2.2: A Dialogue Dataset with Additional Annotation Corrections and State Tracking Baselines. In Proceedings of the 2nd Workshop on Natural Language Processing for Conversational AI (pp. 109-117).


\bibitem{zhang-etal-2020-dialogpt} Zhang, Y., Sun, S., Galley, M., Chen, Y. C., Brockett, C., Gao, X., ... \& Dolan, W. B. (2020, July). DIALOGPT: Large-Scale Generative Pre-training for Conversational Response Generation. In Proceedings of the 58th Annual Meeting of the Association for Computational Linguistics: System Demonstrations (pp. 270-278).

\bibitem{zhang-etal-2023-monet} Zhang, H., Bao, J., Sun, H., Wu, Y., Li, W., Cui, S., \& He, X. (2023). MoNET: Tackle State Momentum via Noise-Enhanced Training for Dialogue State Tracking. In Findings of the Association for Computational Linguistics: ACL 2023 (pp. 520–534). Association for Computational Linguistics.

\bibitem{zhao-etal-2020-knowledge-grounded} Zhao, X., Wu, W., Xu, C., Tao, C., Zhao, D., \& Yan, R. (2020, November). Knowledge-Grounded Dialogue Generation with Pre-trained Language Models. In Proceedings of the 2020 Conference on Empirical Methods in Natural Language Processing (EMNLP) (pp. 3377-3390).

\end{thebibliography}
 \bibliographystyle{abbrv}


\end{document}